\documentclass[journal,twoside,web]{ieeecolor}
\usepackage{lcsys}
\usepackage{cite}
\usepackage{textcomp}
\def\BibTeX{{\rm B\kern-.05em{\sc i\kern-.025em b}\kern-.08em
    T\kern-.1667em\lower.7ex\hbox{E}\kern-.125emX}}
\usepackage{graphicx} 
\usepackage{bbm}
\usepackage[english]{babel}
\usepackage{amsmath,amsfonts,amssymb, amsthm}
\usepackage{graphicx}
\usepackage[colorlinks=true, allcolors=blue]{hyperref}
\usepackage{parskip}
\usepackage{tabularx}
\usepackage{bm}
\usepackage{hhline}
\usepackage{multirow}
\usepackage{algorithm}
\usepackage{algpseudocode}
\usepackage{etoolbox}
\usepackage{subcaption}
\usepackage{algpseudocode}
\usepackage{mathtools}
\usepackage{nth}
\usepackage{diagbox}
\usepackage{float}
\usepackage{tabstackengine}
\usepackage{tablefootnote}
\usepackage{dsfont}

\patchcmd{\thebibliography} 
  {\settowidth}
  {\setlength{\parsep}{0pt}\setlength{\itemsep}{0pt plus 0.1pt}\settowidth}
  {}{}

\newtheorem{theorem}{Theorem}
\newtheorem{proposition}{Proposition}

\newtheorem{remark}{Remark}

\renewcommand{\eqref}[1]{(\ref{#1})}
\renewcommand{\t}{^{\mbox{\tiny\sf T}}} 

\newcommand{\N}{\mathcal{N}}

\definecolor{rho1}{rgb}{0. , 0.976, 0.601}
\definecolor{rho0}{rgb}{0.255, 0.413, 0.885}
\definecolor{rho05}{rgb}{0.125, 0.695, 0.664}

\newcommand{\X}{\mathcal{X}}
\newcommand{\half}{\frac{1}{2}}

\newcommand{\R}{\mathbb{R}}
\renewcommand{\P}{\mathbb{P}}
\newcommand{\E}{\mathbb{E}}
\newcommand{\tr}{\mathrm{tr}}
\renewcommand{\d}{\mathrm{d}}

\newcommand{\dw}{\mathrm{d}w}
\newcommand{\dt}{\mathrm{d}t}

\renewcommand{\baselinestretch}{0.95}

\begin{document}

\title{\huge{Steering Large Agent Populations using Mean-Field Schr\"odinger Bridges with Gaussian Mixture Models}}
\author{George Rapakoulias, \IEEEmembership{Student Member, IEEE}, Ali Reza Pedram,
and Panagiotis Tsiotras, \IEEEmembership{Fellow, IEEE}
\thanks{Support for this work has been provided by 
ONR award N00014-18-1-2828 and NASA ULI award \#80NSSC20M0163.
This article solely reflects the opinions and conclusions of its authors and not of any NASA entity. 
George Rapakoulias acknowledges financial support from the A. Onassis Foundation Scholarship.
}
\thanks{The authors are with the Department of Aerospace Engineering of Georgia Institute of Technology. Corresponding author: \texttt{grap@gatech.edu}}
}

\maketitle
\begin{abstract}
    The Mean-Field Schr\"odinger Bridge (MFSB) problem is an optimization problem aiming to find the minimum effort control policy to drive a McKean-Vlassov stochastic differential equation from one probability measure to another.
    In the context of multi-agent control, the objective is to control the configuration of a swarm of identical, interacting cooperative agents, as captured by the time-varying probability measure of their state.
    Available methods for solving this problem for distributions with continuous support rely either on spatial discretizations of the problem's domain or on approximating optimal solutions using neural networks trained through stochastic optimization schemes.
    For agents following Linear Time Varying dynamics, and for Gaussian Mixture Model boundary distributions, we propose a highly efficient parameterization to approximate the optimal solutions of the corresponding MFSB in closed form, without any learning step.
    Our proposed approach consists of a mixture of elementary policies, each solving a Gaussian-to-Gaussian Covariance Steering problem from the components of the initial mixture to the components of the terminal mixture.
    Leveraging the semidefinite formulation of the Covariance Steering problem, the proposed solver can handle probabilistic constraints on the system's state while maintaining numerical tractability.
    We illustrate our approach on a variety of numerical examples. 
\end{abstract}

\begin{IEEEkeywords}
Mean-field control, Multi-agent control, \\
Schr\"odinger Bridges, Stochastic control.
\end{IEEEkeywords}

\section{Introduction}
\IEEEPARstart{D}esigning optimal control strategies for large multi-agent systems is a difficult optimization problem, mainly due to the well-known curse of dimensionality related to the exponential computational complexity of the problem in the number of agents.
To overcome this problem, assuming homogeneity among the agents, one can view them as a different realization of the same dynamical system, and in the limit of an infinite number of agents, formulate the problem as a control problem in the space of probability distributions \cite{chen2023density}. 
This approach, known as a mean-field control problem \cite{lasry2007mean, huang2006large}, was originally presented as a general tool for studying large-scale systems, with applications ranging from finance and social sciences \cite{carmona2018probabilistic} to engineering and control \cite{chen2018steering}, among many others.
In the context of swarm robotics and multi-agent control, this macroscopic perspective is highly favorable in large-scale problems since its computation does not depend on the number of agents involved, making it a suitable tool for applications such as coverage and surveillance, mapping, and task allocation problems \cite{elamvazhuthi_mean-field_2019, elamvazhuthi2015optimal, elamvazhuthi_controllability_2021}.

To be more precise, consider $N$ identical, interacting, cooperative agents, where $N$ is large. 
Each agent $i \in \{1,2,\dots, N\}$ follows the same stochastic dynamics of the form
\begin{align} \nonumber
 \d x^i_t = \, & -\nabla V_t(x_t^i) \, \dt - \frac{1}{N-1}\textstyle\sum_{j \neq i}   \nabla W_t (x^i_t - x^j_t) \, \dt \\ \label{finp_SDE}
 & + B_t u^i_t \, \d t + D_t \, \dw_t,
\end{align}
where  $V_t$ and $W_t$ are potentials guiding the system’s evolution. 
These terms correspond to the potential energy of a single agent, and the interaction energy between agents, respectively.
The random variables $x_t$ and  $u_t$  lie in $n$- and $m$-dimensional Euclidean spaces $\mathbb{R}^n$ and $\mathbb{R}^m$, respectively, and $\d w_t$ is the $q$-dimensional Brownian motion increment. 

In the specific case where the potentials are quadratic, i.e., $W_t(x) = \half \, x\t Q_t x$ and $  V_t(x) = \half \, x \t P_t x$, \eqref{finp_SDE} simplifies to a system of linear Stochastic Differential Equations (SDEs)
\begin{equation*} 
 \d x_t^i \!=\!-\!\left(P_t \!+\! Q_t \right)\! x^i_t \, \dt  + \frac{1}{N-1} Q_t \! \sum_{j \neq i} \! x^j_t \, \dt + B_t u_t \, \dt + D_t \d w_t. 
\end{equation*}
Instead of controlling each agent individually by assigning a target state, we want to control their collective behavior. In the limit of $N \rightarrow \infty$, this behavior is captured by a time-dependent continuous distribution, denoted $\rho_t(x)$, supported 
on $\R^n$. Furthermore,
\begin{equation} \label{MF_approx}
    \lim_{N\rightarrow \infty} \frac{1}{N-1} \sum_{j\neq i} x^i_t =\!\int x \rho_t(x) \d x = \E_{x \sim \rho_t}[x] \!\triangleq \bar{x}_t.
\end{equation}
With \eqref{MF_approx} in mind, and defining the potentially time-varying matrices $A_t = -P_t -Q_t, \quad \bar{A}_t = Q_t$, a generic agent with state $x_t$ evolves according to the well-known linear McKean-Vlasov SDE (MV-SDE)
\begin{equation} \label{mfSDE}
 \d x_t = A_t x_t \, \dt + \bar{A}_t \bar{x}_{t} \, \dt + B_t u_t \, \d t + D_t \, \d w_t,
\end{equation}
while the density of the state of \eqref{mfSDE}  evolves according to the Fokker Planck Kolmogorov (FPK) equation~\cite{backhoff2020mean}
\begin{equation} \label{mfFPK}
    \frac{\partial \rho_t}{\partial t} + \nabla \cdot \left( (A_t x + \bar{A}_t \bar{x}_{t} + B_t u) \rho_t \right) -\frac{1}{2} \tr( D_t D_t\t \nabla ^2 \rho_t) = 0. 
\end{equation}
The control of the density of \eqref{mfSDE} with prescribed initial and final distributions is known as the Mean Field Schr\"{o}dinger Bridge (MFSB) \cite{backhoff2020mean} problem and reads as follows
\begin{subequations} \label{MFSB}
\begin{align}
\min_{u \in \mathcal{U}} \;&J\triangleq \E \left[ \int_{0}^{1}{ \| u_t(x_t) \|^2 \,
\d t} \right], \label{MFSB_cost} \\
\mathrm{s.t.} \  & \d x_t = A_t x_t \, \d t  + \bar{A} \bar{x}_t \, \d t + B_t u_t \, \d t + D_t\, \d w_t,\label{MFSB_dyn} \\
& x_0 \sim \rho_0, \quad x_1 \sim \rho_1, \label{MFSB_BC}
\end{align}
\end{subequations}
%
where $\rho_0$ and $ \rho_1$ denote the given initial and final state densities, respectively, and where the time horizon is set to $1$ without loss of generality, assuming appropriate rescaling of the system matrices.

Before discussing our solution approach, we present a concise literature review regarding Problem \eqref{MFSB} and its generalizations. 
First, in the absence of the mean-field term $\bar{A}_t$, the problem is the well-studied Schr\"odinger Bridge (SB) problem over a linear dynamical system \cite{chen2016optimal}. 
In the special case of Gaussian boundary distributions $\rho_0, \rho_1$, the problem reduces to the well-known covariance steering problem \cite{chen2015optimal}, which has been studied extensively in the control community in a multitude of different scenarios, for example, under distributional uncertainty \cite{pilipovsky2024dust, renganathan2023distributionally}, for nonlinear systems \cite{kumagai2024sequential} and 
distributed \cite{saravanos2021distributed, saravanos2024distributed} and hierarchical schemes \cite{saravanos2023distributed}, to name a few.

Although covariance steering solvers work primarily with Gaussian boundary distributions, recent works have proposed methods to handle the case of Gaussian Mixture Model (GMM) boundary distributions.
In~\cite{balci2023density}, the authors proposed a randomized feedback policy that can steer an initial GMM to a terminal GMM subject to deterministic discrete-time linear dynamics, while in~\cite{kumagai2024chance} the method was extended to solve constrained problems. 
These methods provide a feasible solution to the problem; however, they sacrifice optimality for computational tractability and work well only for deterministic systems.
In \cite{rapakoulias2024go}, the authors proposed a closed-form deterministic non-linear feedback law to solve the SB problem with continuous-time linear prior dynamics and GMM boundary conditions, using a dynamically weighted sum of elementary policies where each one solves a covariance steering problem from each component of the initial to each component of the terminal Gaussian mixture.
Also, in ~\cite{mei2024flow}, the authors propose a neural-network-based scheme to approximate a feasible policy for the SB problem and arrive at a similar policy to \cite{rapakoulias2024go} for the special case of GMM boundary distributions.

Returning to problems with mean-field terms, for the specific case of unconstrained Linear Time Varying systems with linear mean-field terms, as in equation \eqref{MFSB}, it has been shown in  \cite{chen2018steering} that the problem can be decomposed into two parts: a standard SB problem for a shifted set of zero-mean boundary distributions, which is still challenging to compute, and an easy-to-compute feedforward term to compensate for the mean-field effect. For more general problems with boundary distributions available only through samples, and for non-linear prior dynamics with mean-field terms, solutions are difficult to obtain since the problem consists of a combination of a general optimal transport problem and a non-linear optimal control problem, neither of which can be solved in closed form, except for very few special cases~\cite{terpin2024dynamic}.
Notable works addressing the mean-field SB in this more general setting, under some assumptions on the structure of the dynamical system are \cite{liu2022deep, ruthotto2020machine, lin_alternating_2021}, which utilize neural networks to approximate the optimal control policies, \cite{caluya2021wasserstein}, which uses a proximal recursion scheme leveraging the Jordan-Kinderleher-Otto (JKO) scheme and evaluates the optimal solution at the finite number of samples rather than the entire space, and~\cite{chen2023density} that discretizes the state space and formulates the problem as a multi-marginal discrete optimal transport problem.

As evident from the above discussion, the available methods in the literature for solving Problem \eqref{MFSB} can be categorized into three types:  Covariance steering type methods that are computationally efficient but are limited to Gaussian boundary distributions, domain-discretization-based methods, which although accurate, do not scale well to high-dimensional problems, and methods utilizing neural networks, which despite being general, are computationally very demanding and lack performance guarantees. 
In this context, we propose a new method that maintains the numerical tractability and accuracy of covariance steering algorithms, but handles complicated boundary distributions using mixture models.
More specifically, in this paper, we present the following main contributions:

\begin{itemize}
    \item Building on the decomposition technique introduced in \cite{chen2018steering}, we extend the results of \cite{rapakoulias2024go} to solve the MFSB  problem. 
    Our method splits \eqref{MFSB} into a deterministic mean steering problem and an SB problem, with the latter solved via the mixture-based approach from \cite{rapakoulias2024go}.
    
    \item  
    We then introduce a more general problem formulation that allows us to solve SB/MFSB problems with probabilistic state constraints by leveraging the semidefinite formulation of the covariance steering problem subject to chance constraints.

    \item
    We use our proposed algorithm to solve the path planning problem for stochastic multi-agent systems in the limit of infinite agents.
\end{itemize}

\section{Preliminaries}

\subsection{The Gaussian Schr\"odinger Bridge} \label{OCS}

The SB problem with Gaussian marginals, known as the Optimal Covariance Steering (OCS) problem in the control community \cite{chen2015optimal}, has been extensively studied in the literature and can be solved either analytically for simple choices of prior dynamics \cite{bunne2023Schrodinger} or as a convex 
semidefinite optimization problem for general linear dynamical systems both for continuous and discrete time cases \cite{chen2015optimal2, liu2022optimal}.
Because we use the OCS solution as a building block to construct a policy that works with general GMM boundary distributions, we briefly review the available methods for its solution here.

To this end, consider the optimization problem
\begin{subequations} \label{GSB}
\begin{align}
\min_{u \in \mathcal{U}} \;\;& \E \left[ \int_{0}^{1}{ \| u_t(x_t) \|^2 \, \d t} \right], \label{GSB:cost} \\
\mathrm{s.t.} \; \; & \d x_t = A_t x_t \, \d t + B_t u_t \, \d t + D_t\, \d w_t,\label{GSB:dyn} \\
& x_0 \sim \N(\hat{\mu}_0, \hat{\Sigma}_0), \quad x_1 \sim \N(\hat{\mu}_1, \hat{\Sigma}_1), \label{GSB:BC}
\end{align}
\end{subequations}
where $\hat{\mu}_0, \hat{\Sigma}_0, \hat{\mu}_1, \hat{\Sigma}_1$ are the means and covariances of the initial and final Gaussian boundary distributions, respectively. 
Since the optimal distribution of the state that solves \eqref{GSB} remains Gaussian \cite{chen2015optimal, liu2022optimal}, i.e., $x_t \sim \mathcal{N}(\mu_t, \Sigma_t)$, Problem \eqref{GSB} reduces to that of the control of the mean and the covariance of the state. 
It can be shown \cite{liu2022optimal, chen2016optimal} that the optimal policy solving \eqref{GSB} has the form 
\begin{equation} \label{feedback}
    u_t(x) = K_t(x-\mu_t) + v_t.
\end{equation}
Here, $K_t \in \mathbb{R}^{m\times n}$
is a time-varying feedback gain matrix, and $v_t \in \mathbb{R}^m$ is a feedforward control term. The vector $\mu_t \in \mathbb{R}^m$ denotes the mean of the state distribution at time $t$.
Substituting \eqref{feedback} in \eqref{GSB:dyn}, performing the change of variables $U_t = K_t \Sigma_t$ and introducing the new variable $Y_t = U_t \Sigma_t U_t\t$, Problem \eqref{GSB} becomes 
\begin{subequations} \label{GSB_mom}
\begin{align}
\min \;\; & \E \left[ \int_{0}^{1}{ \tr(Y_t) + v_t\t v_t  \, \d t} \right], \label{GSB_mom:cost} \\
\mathrm{s.t.} \; \;  & Y_t = U_t \Sigma_t U_t\t, \label{GSB_mom:relax} \\
& \dot{\Sigma}_t = A_t \Sigma_t + \Sigma_t A_t\t + B_t U_t + U_t\t B_t\t + D_t D_t\t,\label{GSB_mom:cov} \\
& \dot{\mu}_t = A_t \mu_t + B_t v_t, \label{GSB_mom:mean} \\
& \mu_0 = \hat{\mu}_0, \, \Sigma_0 = \hat{\Sigma}_0, \mu_1 = \hat{\mu}_1, \, \Sigma_1 = \hat{\Sigma}_1, \label{GSB_mom:BC1}
\end{align}
\end{subequations}
where variables are $U_t, \Sigma_t, Y_t, v_t$, and $ \mu_t$.
Relaxing \eqref{GSB_mom:relax} to a semidefinite inequality and using the Schur complement Lemma 
yields ~\cite{chen2015optimal2}  
\begin{equation}
\begin{bmatrix} 
\Sigma_k & U_k\t \\
U_k & Y_k
\end{bmatrix} \succeq 0,
\end{equation} 
and turns \eqref{GSB_mom} into a semidefinite program (SDP), which can be discretized in time and solved efficiently with any off-the-shelf SDP optimizer.

The semidefinite formulation \eqref{GSB_mom} offers the flexibility of adding probabilistic constraints on the state of the dynamical system \eqref{GSB:dyn}.
However, because the problem is formulated in continuous time, imposing constraints on the entire trajectory is difficult.
To this end, in a similar setting to the discrete-time chance constraint formulation \cite{rapakoulias2023discrete}, we may constrain the probability that the state belongs to a safe set at a time $t$, simply by imposing constraints on the moments $\mu_t, \Sigma_t$. 
\vspace{-2mm}
\subsection{The Schr\"odinger Bridge Problem with GMM boundary distributions} \label{GMMflow}
Before solving the MFSB Problem \eqref{MFSB},
we discuss the solution of the corresponding SB problem.
Although an exact, globally optimal solution to this problem is difficult to obtain when the boundary distributions are GMMs, approximate solutions can be achieved with the technique proposed in \cite{rapakoulias2024go}.
In this section, we review this technique, since it serves as the backbone of our analysis.
To this end, consider a generalization of \eqref{GSB} where the boundary distributions are now Gaussian mixture models, that is, 
\begin{subequations} \label{GMM_SB}
\begin{align}
& \min_{u \in \mathcal{U}} \; J_{\mathrm{GMM}} \triangleq \E \left[ \int_{0}^{1}{ \| u_t(x_t) \|^2 \,
\d t} \right], \label{GMM_SB:cost} \\
& \d x = A_t x_t \, \d t + B_t u_t \, \d t + D_t\, \d w, \label{GMM_SB:dyn} \\
& x_0 \!\sim\!  \textstyle\sum_{i=1}^{N_0}  \alpha_0^i \N(\mu^i_0, \Sigma^i_0), \, x_1 \!\sim\!  \textstyle\sum_{j=1}^{N_1} \alpha_1^j \N( \mu^j_1, \Sigma^j_1).  \label{GMM_SB:BC1}
\end{align}
\end{subequations}
%
%
%
A set of feasible solutions to \eqref{GMM_SB} can be constructed using the following theorem.

\begin{theorem} [\cite{rapakoulias2024go} Theorem 1]\label{feasibility_thm}
    Consider Problem \eqref{GMM_SB}, with $N_0$ components in the initial mixture and $N_1$ components in the terminal mixture. 
    Assume that $u_{t|ij}$ is the conditional policy that solves the $(i,j)$-OCS problem, that is, a policy of the form \eqref{feedback} that steers the system \eqref{GMM_SB:dyn} from the $i$-th component of the initial mixture to the $j$-th component of the terminal mixture, and let the resulting probability flow be $\rho_{t|ij} = \N(\mu_{t|ij}, \Sigma_{t|ij})$. 
    Furthermore, let $\lambda_{ij} \geq 0$ such that, for all $j\in\{1, 2, \dots, N_1\}$, $\sum_i \lambda_{ij} = \alpha_1^j$ and for all $i \in\{1, 2, \dots, N_0\}$, $\sum_{j} \lambda_{ij} = \alpha_0^i$.
    Then, the policy
    \begin{equation}\label{lambda_pol}
        u_t(x) =  \textstyle\sum_{i,j} {u_{t|ij}(x) \frac{ \rho_{t|ij}(x)\lambda_{ij}}{\sum_{i,j} \rho_{t|ij}(x) \lambda_{ij}}},
    \end{equation}
    is a feasible policy for Problem \eqref{GMM_SB}, and
    the corresponding probability flow is
\vspace{-2mm}
\begin{equation}\label{rho_lambda}
    \rho_t(x) =  \textstyle\sum_{i,j} \rho_{t|ij}(x) \lambda_{ij}.
\end{equation}
\end{theorem}
\vspace{-2mm}
The mixture policy \eqref{lambda_pol} is a weighted average of conditional policies, weighted according to  $\lambda_{ij} \rho_{t|ij}(x)$, while the denominator $\sum_{i,j} \rho_{t|ij}(x) \lambda_{ij}$ is just a normalizing constant.
Since $\rho_{t|ij}(x)$ is a Gaussian distribution centered at the mean of the $(i,j)$-OCS flow at time $t$, this weighting scheme prioritizes the conditional policies whose mean is closer to the value of $x$ at the time~$t$.
\vspace{-1mm}

Finding the optimal policy within the feasible set defined by Theorem \ref{feasibility_thm} is challenging. 
To this end, instead of optimizing \eqref{GMM_SB:cost}
directly, 
we derive an upper bound, which is linear with respect to the transport plan $\lambda_{ij}$, and is therefore easier to optimize.
More formally, we have the following theorem.
\begin{theorem} (\cite[Theorem 2]{rapakoulias2024go}) \label{OT_thm}
Let $J_{ij}$ be the optimal cost of the $(i,j)$-OCS subproblem with marginal distributions given by the i-th component of the initial mixture and the j-th component of the terminal mixture.
Then, the cost function of the linear optimization problem
\begin{subequations}\label{fleet_OT}
\begin{align}
& \min_{\lambda_{ij}} \quad  J_{\mathrm{OT}} \triangleq  \textstyle\sum_{i,j} \lambda_{ij} J_{ij}, \label{OTcost} \\
& \textup{such that for all $ i = 1, 2, \dots, N_0, \;  j = 1, 2, \dots, N_1 $} \nonumber \\
&  \textstyle\sum_j \lambda_{ij} = \alpha_0^i ,  \textstyle\sum_i \lambda_{ij}= \alpha_1^j, \lambda_{ij} \geq 0,
\label{lambda_f}
\end{align}
\end{subequations}
provides an upper bound for \eqref{GMM_SB:cost}, that is, $J_{\mathrm{GMM}} \leq J_{\mathrm{OT}}$, for all values of $\lambda_{ij}$ satisfying \eqref{lambda_f}.
\end{theorem}
\vspace{-2mm}
\begin{proof}
    An immediate proof follows from the discrete version of Jensen's inequality, as shown in \cite{rapakoulias2024go}. Here, we provide an alternative proof using the variance decomposition identity, which allows us to quantify the gap of the upper bound. 
    Specifically, recall that for any random variable $x \in \R^n$, we have  $\|\E[x]\|^2 = \E[\|x\|^2] - \E[\|x  - \E[x]\|^2 ]$.
    Expressing \eqref{lambda_pol} as $u_t(x) = \E [ \omega_t(x) ]$, where the discrete variable $\omega_t(x)$ is defined by $\omega_t(x) \triangleq \{u_{t|ij}(x) \, \, \mathrm{w.p.} \, \,  \frac{ \rho_{t|ij}(x)\lambda_{ij}}{\rho_t(x)} \}$ we obtain:
\vspace{-2mm}
    \begin{align} \nonumber
     J_{\mathrm{GMM}} \! &=\! \int_0^1 \!\! \int_{\R^n} \rho_{t} \| u_t \|^2 \, \d x \, \d t \\ 
    & =\! J_{\mathrm{OT}} - \!\!\int_0^1  \!\!\int_{\mathbb{R}^n} \textstyle \!\sum_{i,j} \lambda_{ij} \rho_{t|ij} \|u_{t|ij}- u_t\|^2  \, \d x \, \d t. \label{final_bound}
\end{align}
The second term in \eqref{final_bound} is positive by construction, hence dropping it yields the desired result. 
\end{proof}
\vspace{-1mm}
Theorem \ref{OT_thm} allows for a very efficient computation of the SB policy since the conditional policies in Theorem~\ref{feasibility_thm} can be computed independently and are decoupled from the calculation of the transport plan $\lambda_{ij}$, which is computed as the solution of \eqref{fleet_OT}.
Furthermore, although the upper bound in Theorem \ref{OT_thm} introduces some sub-optimality, in practice, we observed that policy \eqref{lambda_pol} results in lower transport costs in SB problems with GMM boundary distributions compared to neural approaches \cite{rapakoulias2024go}, which is attributed to the often imperfect neural network training.
\begin{remark} By substituting the policy \eqref{lambda_pol} in \eqref{final_bound}, it can be shown that when the components of the flow \eqref{rho_lambda} are well-separated, 
the second term in \eqref{final_bound} tends to zero, making the upper bound in Theorem \ref{OT_thm} tight. A more rigorous discussion of this case is differed to Appendix \ref{App:A}.
\end{remark}

\section{Unconstrained Mean Field Sch\"oringer Bridges} \label{unc_MFSB}

In this section, we extend the above approach to solve the corresponding MFSB problem with GMM boundary conditions.
We first study the problem without state constraints, that is, we solve
\begin{subequations} \label{GMM_MFSB}
\begin{align}
&\min_{u \in \mathcal{U}} \; \E \left[ \int_{0}^{1}{ \| u_t(x_t) \|^2 \
\d t} \right], \label{GMM_MFSB:cost} \\
& \d x_t = A_t x_t \, \d t + \bar{A}_t \bar{x}_t \, \d t + B_t u_t \, \d t + D_t\, \d w_t, \label{GMM_MFSB:dyn} \\
& x_0 \!\sim\!  \textstyle\sum_{i=1}^{N_0}  \alpha_0^i \N(\mu^i_0, \Sigma^i_0), \, x_1\!\sim\! \textstyle\sum_{j=1}^{N_1} \alpha_1^j \N( \mu^j_1, \Sigma^j_1), \label{GMM_MFSB:BC1}
\end{align}
\end{subequations}
where $\bar{x}_t = \E[x_t]$.
We follow an approach similar to \cite{chen2018steering}, but tailored to the specific case of the mixture policy defined in Theorem \ref{feasibility_thm}.
Defining $\bar{u}_t = \E[u_t(x_t)]$ and $\tilde{x}_t = x_t - \bar{x}_t, \, \tilde{u}_t = u_t - \bar{u}_t$, and observing that, for all $t \in [0,1]$, $\E[\tilde{x}_t]=0$ and $\E[\tilde{u}_t]=0$, one can rewrite \eqref{GMM_MFSB} as 
\begin{subequations} \label{GMM_MFSB_decoup}
\begin{align}
& \min_{u \in \mathcal{U}} \; \E \left[ \int_{0}^{1}{ \| \tilde{u}_t(x_t) \|^2 + \| \bar{u}_t \|^2 \,
\d t} \right], \label{GMM_MFSB_decoup:cost} \\
& \d \tilde{x}_t = A_t \tilde{x}_t \, \d t + B_t \tilde{u}_t \, \d t + D_t\, \d w, \label{GMM_MFSB_decoup:dev} \\
& \d \bar{x}_t = (A_t + \bar{A}_t) \bar{x}_t \, \d t + B_t \bar{u}_t \d t, \label{GMM_MFSB_decoup:mean} \\
& \E[\tilde{x}_t] = 0, \quad \E[\tilde{u}_t] = 0, \label{GMM_MFSB_remain:zero_mean} \\
& \tilde{x}_0 \sim  \textstyle\sum_{i=1}^{N_0} \alpha_0^i \N(\mu^i_0-\hat{x}_0, \Sigma^i_0),
\label{GMM_MFSB_decoup:BC0}\\
& 
\tilde{x}_1 \sim  \textstyle\sum_{j=1}^{N_1} \alpha_1^j \N(\mu^j_1-\hat{x}_1, \Sigma^j_1), \label{GMM_MFSB_decoup:BC1} \\ \label{GMM_MFSB_decoup:BC2}
& \bar{x}_0 = \hat{x}_0,  \quad  \bar{x}_1 = \hat{x}_1, 
\end{align}
\end{subequations}
where $\hat{x}_0 = \sum_i \mu_0^i w^i, \, \hat{x}_1=\sum_j \mu^j_1 w^j$ are fixed quantities deduced by the parameters of the mixture models representing the means of the boundary distributions \eqref{GMM_MFSB:BC1}.
In \eqref{GMM_MFSB_decoup}, the problem of calculating the mean control $\bar{u}_t$ and the mean state trajectory $\bar{x}_t$ \vspace{-1em} 
\begin{equation}
    \min_{\bar{u}} \; \; \int_{0}^{1}{ \| \bar{u}_t \|^2 \d t} \quad \mathrm{s.t.} \quad  \eqref{GMM_MFSB_decoup:mean}, \, \eqref{GMM_MFSB_decoup:BC2}, \label{GMM_MFSB_mean}
\end{equation}
is decoupled from the rest of the problem, which reads 
\begin{equation} \label{GMM_MFSB_remain}
\min_{\tilde{u_t} \in \mathcal{U}} \; \;  \E \left[ \int_{0}^{1}{ \| \tilde{u}_t(x_t) \|^2 \, \d t} \right] \  \ \mathrm{s.t.}  \ \ \eqref{GMM_MFSB_decoup:dev}, \eqref{GMM_MFSB_remain:zero_mean}- \eqref{GMM_MFSB_decoup:BC1}.
\end{equation}
The solution of \eqref{GMM_MFSB_mean} can be obtained in closed form, while \eqref{GMM_MFSB_remain} is of the form \eqref{GMM_SB} with the additional constraints in~\eqref{GMM_MFSB_remain:zero_mean}.
Next, consider a relaxation of \eqref{GMM_MFSB_remain} where we drop the constraint~\eqref{GMM_MFSB_remain:zero_mean} and solve the remaining problem with the technique of Section \ref{GMMflow}.
Most importantly, we show that for any policy in the feasible set of policies defined in Theorem~\ref{feasibility_thm}, the conditions 
$\E[\tilde{x}_t] = 0, \, \E[\tilde{u}_t] = 0$ will hold automatically due to the structure of the optimal solution of the $(i, j)$-OCS and the fact that the boundary distributions in \eqref{GMM_MFSB_decoup:BC0} and \eqref{GMM_MFSB_decoup:BC1} have zero-mean.
We formalize this in the following proposition. 
\begin{proposition} \label{zero_mean_prop}
    Consider Problem \eqref{GMM_MFSB_remain} without the constraints \eqref{GMM_MFSB_remain:zero_mean}, and let 
   $\tilde{u}_t, \tilde{\rho}_t$ be  a feasible solution pair of the form \eqref{lambda_pol}, \eqref{rho_lambda} for some $\lambda_{ij}$ satisfying the conditions of 
   Theorem~\ref{feasibility_thm}. Then, for any such feasible pair, it holds that $\E[\tilde{x}_t] = 0, \, \E[\tilde{u}_t] = 0 \,\, \forall \,\, t \in [0,1]$. 
\end{proposition}
\vspace{-1.0em}
\begin{proof}
    First,  taking the expectations of Equations \eqref{lambda_pol}, \eqref{rho_lambda}, for the boundary distributions   
    \eqref{GMM_MFSB_decoup:BC0}, \eqref{GMM_MFSB_decoup:BC1}, we get
    \begin{subequations} \label{zero_mean_cond}
    \begin{align}
        & \E[\tilde{x}_t] = \textstyle\sum_{ij} \lambda_{ij} \E[\tilde{x}_{t|ij}] =  \textstyle\sum_{ij} \lambda_{ij} \tilde{\mu}_{t|ij}, \\
        & \E[\tilde{u}_t] =  \textstyle\sum_{ij} \lambda_{ij} \E[\tilde{u}_{t|ij}] =  \textstyle\sum_{ij} \lambda_{ij} \tilde{v}_{t|ij},
    \end{align}
    \end{subequations}
    where the expectations are taken with respect to the underlying distribution of the random variable in the expectation and $\tilde{\mu}_{t|ij}, \tilde{v}_{t|ij}$ are the mean state and control effort of the $(i, j)$-OCS problem with boundary distributions of the $i$-th component of the initial mixture \eqref{GMM_MFSB_decoup:BC0} and the $j$-th component of the terminal mixture \eqref{GMM_MFSB_decoup:BC1}. 
    Calculating $\tilde{\mu}_{t|ij}, \tilde{v}_{t|ij}$ amounts to a minimum-effort control problem for the linear dynamical system defined by the pair $(A_t, B_t)$ with boundary conditions $\tilde{\mu}_{0|ij}=\mu^i_0-\hat{x}_0$, $\tilde{\mu}_{1|ij}=\mu^j_1-\hat{x}_1$. 
    Their optimal values are~\cite{chen2016optimal}
    \vspace{-5mm}
    \begin{subequations} \label{OCP_sol}
        \begin{align}
             \tilde{\mu}^*_{t|ij} & =  \Phi(t, 1) M(1, t) M^{-1}_{10} \Phi_{10} (\mu_0^i - \hat{\mu}_0)  \nonumber \\
            & + M(t, 0) \Phi(1, t)\t M_{10}^{-1}(\mu_1^j - \hat{\mu}_1), \label{OCP_sol:mu}\\
             \tilde{v}^*_{t|ij} & = B_t\t \Phi(1,t) \t  M^{-1}_{10} (\mu_1^j - \hat{\mu}_1)  \nonumber \\
             & - B_t\t \Phi(1,t) \t  M^{-1}_{10} \Phi_{10} (\mu_0^i - \hat{\mu}_0), \label{OCP_sol:v}
        \end{align}
    \end{subequations}
    where $\Phi(t, s)$ is the state transition matrix of $(A_t, B_t)$ calculated from time $s$ to $t$, $M(t, s)$ is the respective controllability Grammian, and $\Phi_{10} \equiv \Phi(1, 0), \, M_{10} = M(1, 0)$.
    It is easy to verify that substituting \eqref{OCP_sol} to \eqref{zero_mean_cond} indeed yields $\E[\tilde{x}_t] = 0, \, \E[\tilde{u}_t] = 0$ regardless of the choice of $\lambda_{ij}$. 
\end{proof}
\vspace{-1em}
An immediate consequence of Proposition \ref{zero_mean_prop} is that given the solution $\tilde{\rho}_t, \tilde{u}_t$ of Problem \eqref{GMM_MFSB_remain} and the solution $\bar{x}_t, \bar{u}_t$ of Problem \eqref{GMM_MFSB_mean}, a solution for the MFSB Problem \eqref{GMM_MFSB} can be constructed using the formulas $\rho_t(x) = \tilde{\rho}_t(x - \bar{x}_t)$ and $u_t(x) = \tilde{u}_t(x-\bar{x}_t) + \bar{u}_t$. 
%
\vspace{-2mm}
\section{Constrained Mean Field Schr\"odiger Bridges}
In this section, we are interested in further extending the computation of the mixture policy \eqref{lambda_pol} to handle constrained problems of the form 
\vspace{-2mm}
\begin{subequations} \label{conMFSB}
\begin{align}
& \min_{u \in \mathcal{U}} \; \E \left[ \int_{0}^{1}{ \| u_t(x_t) \|^2 \,
\d t} \right], \label{conMFSB:cost} \\
& \eqref{GMM_MFSB:dyn}, \eqref{GMM_MFSB:BC1}, \P(x_t \in \mathcal{X}) \geq 1-\delta_t. \label{conMFSB:CC}
\end{align}
\end{subequations}
where $\mathcal{X} = \bigcup_n \X_n \triangleq \bigcup_n \{x \in \R^n: \alpha_n\t x \leq \beta_n\}$. 
Contrary to the technique in Section \ref{unc_MFSB}, we do not split the state and control action into mean and deviation components, as in the decomposition of Problem \eqref{GMM_MFSB} to \eqref{GMM_MFSB_mean} and \eqref{GMM_MFSB_remain} because the two problems are coupled through the chance constraint \eqref{conMFSB:CC}.
In the same way we combine conditional OCS policies to solve the SB problem for GMM boundary conditions in 
Theorem~\ref{feasibility_thm}, our approach will now be to combine conditional policies, each solving a constrained mean-field OCS problem 
and assemble them together to solve the MFSB problem with GMM boundary conditions.
Similarly to the case without the mean-field term, the policy will have the form \eqref{lambda_pol}, and the resulting probability flow will have the form \eqref{rho_lambda}. 

Before presenting the details for the calculation of the conditional mean-field OCS problem, we reformulate the chance constraint \eqref{conMFSB:CC} in terms of the parameters of the conditional $i$-$j$ policies.
To do that, we assume that, similarly to the standard SB problem, the adopted policy will yield a probability flow of the form \eqref{rho_lambda}.
We will later prove that this is indeed the case. 
To transform the constraint \eqref{conMFSB:CC} to constraints on the moments of the conditional probabilities $\rho_{t|ij}=\N(\mu_{t|ij}, \Sigma_{t|ij})$, we allocate the total constraint violation budget $\delta_{t}$ to the probability of the $i$-$j$ component to exit the half-spaces $\X_n$, denoted $\delta_{t|ijn}$.
Using the union bound \cite{kumagai2024chance}, \eqref{conMFSB:CC} can be conservatively replaced by 
    \begin{align} \label{CC_UB}
      & \P(x_{t|ij} \in \X_n)\! \geq\! 1-\delta_{t|ijn}, \;  \text{and} \; \textstyle\sum_{i,j,n} \lambda_{ij} \delta_{t|ijn} \leq \!\delta_t.
    \end{align}
%
The simplest option for implementing \eqref{CC_UB} in practice is to allocate $\delta_t$ equally on all components and half-spaces, selecting $\delta_{t|ijn}=\delta_{t}/N_c$. If this is too conservative for a given application, the results could be refined with an iterative allocation scheme such as the one proposed in \cite{pilipovsky2021covariance, kumagai2024chance}. 
Furthermore, since $x_{t|ij} \sim \N(\mu_{t|ij}, \Sigma_{t|ij})$, the constraint $\P(x_{t|ij} \in \X_n) \geq 1-\delta_{t|ijn}$ is equivalent to 
\begin{equation} \label{sqrt_CC}
     \Phi^{-1}(1-\delta_{t|ijn}) \sqrt{a_n\t \Sigma_{t|ij} a_n} + a_n\t\mu_{t|ij} \leq \beta_n,
\end{equation}
where $\Phi^{-1}(\cdot)$ is the inverse cumulative distribution function for the normal distribution.
Following the technique in \cite{rapakoulias2023discrete}, the constraint \eqref{sqrt_CC} is linearized around some reference values of the $(i,j)$ problem, yielding a linear constraint of the form
\begin{equation} \label{linear_CC}
    \ell_{ijn} \t \Sigma_{t|ij} \ell_{ijn} + \mu_{t|ij} \t a_{ijn} + b_{ijn} \leq 0,
\end{equation}
where $b_{ijn} = -\beta_n - \Phi^{-1}(1-\delta_{t|ijn}) \frac{1}{2} \sqrt{ a_{n}\t \Sigma_r a_n} $ and
$\ell_{ijn} = \sqrt{\Phi^{-1}(1-\delta_{t|ijn}) \frac{1}{2 \sqrt{ a_n \t \Sigma_r a_n}}} a_n$, and $\Sigma_r$ being a reference value used for the linearization. 



We now present the $(i,j)$ constrained mean-field OCS problem, which reads
\vspace{-1mm}
\begin{subequations} \label{cond_conMFSB} 
\begin{align}
& \min_{u_{t|ij} \in \mathcal{U}} \; \E \left[ \int_{0}^{1}{ \| u_{t|ij}(x_{t|ij}) \|^2 \,
\d t} \right], \label{cond_conMFSB:cost} \\
& \d x_{t|ij} =\! A_t x_{t|ij} \, \d t +\! \bar{A}_t \bar{x}_t \, \d t +\! B_t u_{t|ij} \, \d t +\! D_t\, \d w, \label{cond_conMFSB:dyn} \\
& x_0 \sim \N(\mu^i_0, \Sigma^i), \, x_1 \sim \N( \mu^j_1, \Sigma^j_1), \\
& \P(x_{t|ij} \in \mathcal{X}) \geq 1-\delta_{t|ij}. \label{cond_conMFSB:CC}
\end{align}
\end{subequations}
Although solving \eqref{cond_conMFSB} requires the unknown
mean-field term $\bar{x}_t$, 
we provide the following result, which allows us to solve all the mean-field OCS policies jointly.
\begin{theorem} \label{feasibility_MFSB}
Given the optimal solution $u_{t|ij}$ and  $\rho_{t|ij} = \N(\mu_{t|ij}, \Sigma_{t|ij})$ to \eqref{cond_conMFSB} and a transport plan $\lambda_{ij}$ satisfying the conditions of Theorem \ref{feasibility_thm}, 
the mixture policy 
\eqref{lambda_pol} is a feasible policy for \eqref{cond_conMFSB}, and the resulting probability flow is \eqref{rho_lambda}.
\end{theorem}
\begin{proof}
    The proof follows from the fact that when the conditional policies solve \eqref{cond_conMFSB}, the pair of \eqref{lambda_pol} and  \eqref{rho_lambda} satisfies the mean-field FPK Equation \eqref{mfFPK}. Due to its similarity with the proof of Theorem~\ref{feasibility_thm}, the details are omitted. 
\end{proof}
\vspace{-1.0em}

With \eqref{feasibility_MFSB} in mind, taking the expectation of \eqref{rho_lambda}, the mean-field term $\bar{x}_t$ can be calculated explicitly, yielding $\bar{x}_t = \sum_{ij} \lambda_{ij} \mu_{t|ij},$
which suggests a coupling between the Gaussian mean-field OCS problems.
Using Theorems \ref{feasibility_MFSB} and \ref{OT_thm}, we can assemble all the conditional mean-field OCS problems into a single coupled 
optimization problem using the moment version of \eqref{cond_conMFSB}, which yields 
\begin{subequations} \label{BIG_SDP}
\begin{align}
& \min \;  \textstyle\sum_{ij} \lambda_{ij} \int_{0}^{1}{ \tr (Y_{t|ij}) +  \| v_{t|ij} \|^2 \, \d t}, \label{BIG_SDP:cost} \\
& \text{such that for all $ i = 1, 2, \dots, N_0, \;  j = 1, 2, \dots, N_1 $} \nonumber \\ 
& \begin{bmatrix} 
\Sigma_{t|ij} & U_{t|ij}\t \\
U_{t|ij} & Y_{t|ij}
\end{bmatrix} \succeq 0, \label{SDP_con} \\
& \dot{\Sigma}_{t|ij} \!=\! A_t \Sigma_{t|ij} \!+\! \Sigma_{t|ij} A_t\t \!+\! B_t U_{t|ij}\!+\! U_{t|ij}\t B_t \t \!+\! D_t D_t\t, \label{cov_prog}\\
& \dot{\mu}_{t|ij} = A \mu_{t|ij} + \bar{A} \left( \textstyle\sum_{ij} \lambda_{ij} \mu_{t|ij} \right)+ B v_{t|ij}, \label{BIG_SDP:mean_prop} \\
& \mathrm{\eqref{lambda_f}}, \mu_{0|ij} \!=\! \mu^i_0, \, \Sigma_{0|ij} \!=\!\Sigma_0^i, \, \mu_{1|ij} \!=\! \mu^j_{1}, \, \Sigma_{1|ij}\!=\!\Sigma_1^j,  \\
& \ell_{ijn} \t \Sigma_{t|ij} \ell_{ijn} + \mu_{t|ij} \t a_{ijn} + b_{ijn} \leq 0, 
%
\end{align}
\end{subequations}
Problem \eqref{BIG_SDP} is a bilinear semidefinite program due to the products of the mixing weights $\lambda_{ij}$ and the policy parameters of the $(i,j)$ subproblem.
To solve this problem, we alternate between fixing the transport plan $\lambda_{ij}$ and solving for the conditional OCS parameters, and keeping the conditional policy parameters fixed and optimizing for $\lambda_{ij}$ by solving \eqref{fleet_OT} until the iteration converges. 
Terminating the iteration with an update in the $(i,j)$ policy parameters, i.e., solving \eqref{BIG_SDP} for a fixed $\lambda_{ij}$ ensures that the solution will be feasible, although the transport plan $\lambda_{ij}$ might be sub-optimal.
In practice, we observed that a few iterations of this scheme are enough for convergence in all problems we solved. 
\vspace{-2mm}
\section{Numerical examples}
\begin{figure*}
\centering
\begin{subfigure}{.1902\textwidth}
  \centering
  \includegraphics[width=1\linewidth]{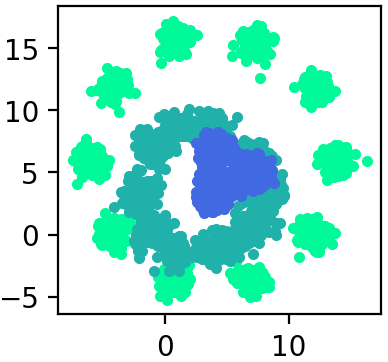} 
  \caption{}
  \label{fig:1a}
\end{subfigure}
\begin{subfigure}{.167\textwidth}
  \centering
  \includegraphics[width=1 \linewidth]{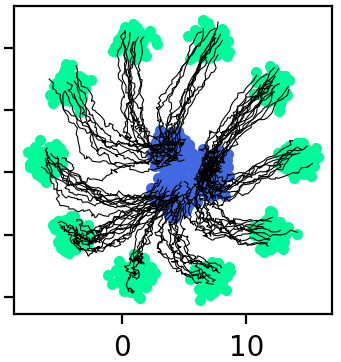}
  \caption{}
  \label{fig:1b}
\end{subfigure}
\hfill
\begin{subfigure}{.221\textwidth}
  \centering
  \includegraphics[width=1 \linewidth]{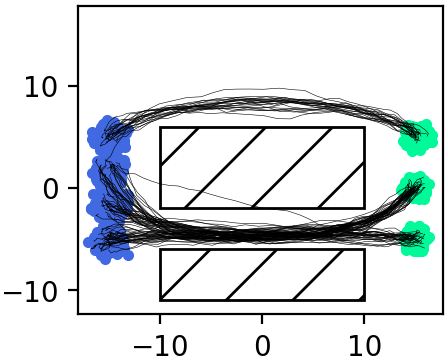}
  \caption{}
  \label{fig:1c}
\end{subfigure}
\begin{subfigure}{.185\textwidth}
  \centering
  \includegraphics[width=1 \linewidth]{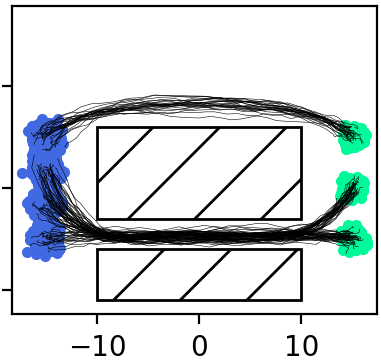}
  \caption{}
  \label{fig:1d}
\end{subfigure}
\begin{subfigure}{.184 \textwidth}
  \centering
  \includegraphics[width=1 \linewidth, trim=2 0 0 0,clip]{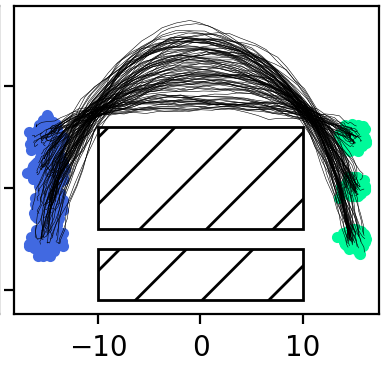}
  \caption{}
  \label{fig:1e}
\end{subfigure}
\caption{(a): \textcolor{rho0}{$\rho_0$}, \textcolor{rho05}{$\rho_{0.5}$}, and \textcolor{rho1}{$\rho_1$} for Problem 1. (b): 100 agent trajectories for Problem 1. (c)-(e): \textcolor{rho0}{$\rho_0$} to \textcolor{rho1}{$\rho_1$} for Problem 2 overlaid with 100 agent trajectories. The shaded regions represent state constraints that the trajectories must avoid.}
\vspace{-4mm}
\end{figure*}

\textbf{Problem 1:} We first apply the algorithm in Section \ref{unc_MFSB} to solve an unconstrained MFSB problem.
We use the LTI dynamical system of the form \eqref{MFSB_dyn} with 
$A = \bar{A} = B = D =I_2$, 
and for boundary distributions the GMMs of Fig.~\ref{fig:1a}-\ref{fig:1b}.
We use blue for the initial and light green for the final distributions.
We discretize the continuous time problem in $101$ time steps and solve the resulting SDP problem with MOSEK \cite{aps2020mosek}. 
The resulting agent trajectories are illustrated in Fig.~\ref{fig:1b}.
We note that for all our simulations, we use the empirical mean-field term computed from  $10,000$ agents.
%
%
%
%
\begin{table}
    \centering
    \caption{Results for Problem 2.\vspace{-0.5em}}
    \begin{tabular}{|l|c|c|c|}
    \hline
         Gap width                              & Fig.\ref{fig:1c} & Fig.\ref{fig:1d} & Fig.\ref{fig:1e}  \\
         \hline
         Total Cost $J$                         & 153.89 & 179.56 & 504.83\\
         \hline
         Cost Upper Bound \eqref{BIG_SDP:cost}  & 306.08 & 357.16 & 999.55\\
         \hline
         $\max_t \P(x_t \notin \mathcal{X})$    & $0.3 \%$ & $0.4 \%$ & $0.5 \%$\\
         \hline
    \end{tabular}
    \label{tab:results}
    \vspace{-1em}
\end{table}


\textbf{Problem 2:} To test the algorithm on a more realistic problem, we considered a swarm of agents following double integrator dynamics, interacting through a negative quadratic potential that incentivizes them to spread apart.
The distribution of the initial and final configurations of the agents is depicted in Fig.~\ref{fig:1c}-\ref{fig:1e}.
The task for this problem is to guide the swarm through an obstacle by either going through a narrow passage or by going around it.
The agents follow the dynamics \eqref{MFSB_dyn} with parameters
\begin{equation} \label{4Dsys}
    A \!=\!\! \begin{bmatrix} 0_2 & I_2 \\ I_2 & 0_2 \end{bmatrix}, \,  B\!=\!\!\begin{bmatrix}0_2 \\ I_2\end{bmatrix}, \, \bar{A} \!=\!\!\begin{bmatrix} 0_2 & 0_2 \\ -I_2 & 0_2 \end{bmatrix}, \, D \!=\! I_4 \vspace{-1em}
\end{equation}

The violation probability for each wall in the obstacles is set to $\delta_{t|ijn}=0.3\%$, giving a total violation of up to $0.9\%$. 
For each conditional policy connecting a component of the initial mixture to a component of the terminal mixture, we solve the problem through both passages. The transport plan in this case becomes a tensor with three indices; however, the corresponding Problem \eqref{fleet_OT} remains an easy-to-solve, low-dimensional linear program.
We solve the problem for three different widths of the passage shown in Fig.~\ref{fig:1c}-\ref{fig:1e}.  
As expected, for a wider passage, the solver routes some of the modes through it, while for smaller passages, the go-around route is preferred.
In all problems, the transport plan converges within a few iterations.
For each run, we report the total resulting true cost of policy \eqref{lambda_pol}, the cost upper bound, which is the value of the optimal cost function \eqref{BIG_SDP:cost}, as well as the maximum constraint violation in Table \ref{tab:results}.
As in our first example, we note that during simulation, we use the empirical mean-field term computed by simulating the interactions of $10,000$ agents. 
%

\vspace{-2mm}
%
\section{Conclusions}
In this paper, we present a method for approximating the solutions of the Mean-Field Schr\"odinger Bridge problem for systems with Linear Time-Varying dynamics and Gaussian mixture model boundary distributions. 
Our approach is based on a mixture policy consisting of linear feedback laws, each solving a conditional problem from one component of the initial mixture to another component of the terminal mixture. 
The elementary policies are weighted with an optimal component-level transport plan. 
In the case of unconstrained problems, the calculation of the transport plan is decoupled from the rest of the problem, and amounts to solving a linear program. 
For the case where the problems contain state constraints, the calculation of the conditional policies and the transport plan are coupled through the chance constraints and the mean-field term.
An iterative optimization method is proposed to solve the resulting optimization problem.  
Although we do not have a proof of convergence, we observed in practice that the method converges in only a few iterations in all numerical examples. 
%


\vspace{-3mm}
\bibliographystyle{ieeetr}
\bibliography{refs}

\appendix
\subsection{Optimality of the mixture policy} \label{App:A}
In this section we rigorously study the gap in the upper bound of Theorem \ref{OT_thm} and show that it asymptotically converges to zero as the components become well-separated.
To this end let $\rho_{t|ij}, u_{t|ij}$ be the solution of the $(i,j)$ Covariance steering problem of the form \eqref{GSB}, with boundary distributions the $i$-th component of the initial and the $j$-th component of the terminal mixture, and let $\rho_t = \sum_{ij} \lambda_{ij} \rho_{t|ij}$ and 
\begin{equation} \label{u_exp}
u_t(x) = \sum_{i,j} {u_{t|ij}(x) \frac{ \rho_{t|ij}(x)\lambda_{ij}} {\rho_{t}(x)}} = \E [ \omega_t(x)],
\end{equation}
where $\omega_{t}(x)$ follows a discrete distribution defined by $\{ \omega_t(x) = u_{t|ij}(x) \, \, \mathrm{w.p.} \, \, \frac{ \rho_{t|ij}(x)\lambda_{ij}} {\rho_{t}(x)} \}$.

Note that for any random variable $x \in \R^n$, the variance decomposition yields 
\begin{equation} \label{var_dec}
    \|\E[x]\|^2 = \E[\|x\|^2] - \E[\|x  - \E[x]\|^2 ].
\end{equation}
Using \eqref{var_dec} and the expression \eqref{u_exp} in \eqref{GMM_SB:cost} we obtain
\begin{align}
    J_{\mathrm{GMM}} & = \int_0^1 \!\! \int_{\R^n} \!\! \rho_{t} (x) \| u_t(x) \|^2 \, \d x \, \d t \\ 
    & =\! J_{\mathrm{OT}} - \!\!\int_0^1  \!\!\int_{\mathbb{R}^n} \textstyle \!\sum_{i,j} \lambda_{ij} \rho_{t|ij} \|u_{t|ij} \!- \!u_t\|^2  \, \d x \, \d t. \label{final_boundA}
\end{align}
The fact that the second term in \eqref{final_boundA} is positive justifies the upper bound in Theorem \ref{OT_thm}. Now, we show that when the conditional densities are well separated, the second term in \eqref{final_boundA} goes to zero, and the bound becomes tight.

First, we note two useful identities that we use to calculate the density of the product and quotient of Gaussian densities.
To this end, consider two Gaussian densities, $\rho_1 = \N(\mu_1, \Sigma_1), \, \rho_2 = \N(\mu_2, \Sigma_2)$. Then
\begin{align}
    &\rho_1 \rho_2 = c_p \N ( \mu_p, \Sigma_p), \label{prod}\\
    &\frac{\rho_1}{ \rho_2} = c_q \N ( \mu_q, \Sigma_q), \label{quotient} 
\end{align}
where 
\begin{align}
    & \mu_p =  (\Sigma_1^{-1} + \Sigma_2^{-1})^{-1}(\Sigma_1^{-1} \mu_1 + \Sigma_2^{-1} \mu_2), \label{prod_mu}\\
    & \Sigma_p = (\Sigma_1^{-1} + \Sigma_2^{-1})^{-1}, \label{prod_sigma} \\
    & c_p = \N(\mu_1 ; \mu_2, \Sigma_1 + \Sigma_2), \\
    & \mu_q = (\Sigma_1^{-1} - \Sigma_2^{-1})^{-1}(\Sigma_1^{-1}\mu_1 - \Sigma_2^{-1} \mu_2), \\
    & \Sigma_q = (\Sigma_1^{-1} - \Sigma_2^{-1})^{-1}, \\
    & c_q = \frac{|\Sigma_2|}{|\Sigma_2 - \Sigma_1|} \frac{1}{\N(\mu_1 ; \mu_2, \Sigma_2 - \Sigma_1)}.
\end{align}

Using \eqref{prod}, it is easy to show that when the components $\rho_{t|ij}$ are well separated, i.e., $\| \mu_{t|ij} - \mu_{t|i'j'} \| > c \,\, \forall \, ,t \in [0,1] $ for some sufficiently large constant $c$, whenever $(i, j) \neq (i', j')$ :
\begin{align}
    & \quad \quad \rho_{t|ij}(x) \rho_{t|i'j'}(x) \leq c_1 \N(\mu, \Sigma) \quad \forall \, \, x \in \R^n  \label{small_1}
\end{align}
for some $\mu, \Sigma$ defined through \eqref{prod_mu}, \eqref{prod_sigma}, and $c_1\rightarrow0$ exponentially as $c \rightarrow \infty$.

With the above tools, expanding the term inside the norm in Equation \eqref{final_bound}, and dropping the dependence on $x$ for notational convenience, we obtain

\begin{align}
    & \, \, \quad \|u_{t|i,j}- u_t\|^2  \\
    & = \bigg\|u_{t|i,j}-  \sum_{i'j'} u_{t|i'j'} \frac{\lambda_{i'j'} \rho_{t|i'j'}}{\rho_t} \bigg\|^2 \\
    & = \bigg\| \frac{ u_{t|ij} \rho_{t}- \sum_{i'j'} u_{t|i'j'} \lambda_{i'j'} \rho_{t|i'j'} }{ \rho_t} \bigg\|^2 \\ 
    & = \Bigg\|\frac{\sum_{i' \neq i, j' \neq j} (u_{t|ij} - u_{t|i'j'}) \lambda_{i'j'} \rho_{t|i'j'}}{ \rho_t}  \Bigg\|^2 \\ 
    &  \leq c_2  \sum_{\substack{i' \neq i \\ j' \neq j}} \big\|u_{t|ij} - u_{t|i'j'} \big\|^2 \left( \frac{ \lambda_{i'j'} \rho_{t|i'j'} }{  \rho_t} \right)^2 \\ 
    &  \leq c_2 \sum_{\substack{ i' \neq i \\ j' \neq j}} \big\|u_{t|ij} - u_{t|i'j'} \big\|^2 \frac{ \lambda_{i'j'} \rho_{t|i'j'} }{  \rho_t} \\ 
    &  \leq c_2  \sum_{\substack{ i' \neq i \\ j' \neq j }} \big\|u_{t|ij} - u_{t|i'j'} \big\|^2 \frac{ \lambda_{i'j'} \rho_{t|i'j'} }{  \max_{i,j} \{ \lambda_{ij} \rho_{t|ij} \} } \label{upper_boundA}
\end{align}

where the $\max$ in \eqref{upper_boundA} selects the component with the largest covariance and $c_2 = (N_0-1)(N_1-1)$, i.e., the cardinality of the sum of policies.
Rearranging terms in Equation \eqref{final_boundA} and substituting the bound \eqref{upper_boundA}, we obtain  
\begin{align}
    0 & \leq J_{\mathrm{OT}} - J_{\mathrm{GMM}} \nonumber \\
    & \leq \int_0^1 \int_{\mathbb{R}^n} \sum_{i,j} \lambda_{ij} \rho_{t|i,j} \|u_{t|i,j}(x)- u_t(x)\|^2  \, \d x \, \d t \nonumber \\ 
    & \leq c_2 \!\! \int_0^1 \!\! \int_{\mathbb{R}^n} \sum_{i,j} \!\sum_{\substack{  i' \neq i \\ j' \neq j} } \big\|u_{t|ij} \! - \! u_{t|i'j'} \big\|^2 \frac{ \lambda_{i'j'} \lambda_{ij} \rho_{t|i'j'} \rho_{t|i,j}  }{  \max_{i,j} \{ \lambda_{ij} \rho_{t|ij} \} }  \, \d x \, \d t \nonumber \\
    %
    %
    & =   \! c_2 \, c_1 \!\! \int_0^1 \! \int_{\mathbb{R}^n} \! \sum_{i,j} \!  \sum_{ \substack{i' \neq i \\ j' \neq j } }  \! \big\|u_{t|ij} \!-\! u_{t|i'j'} \big\|^2 \frac{ \lambda_{i'j'} \lambda_{ij} \N(\mu, \Sigma) }{  \max_{i,j} \{ \lambda_{ij} \rho_{t|ij} \} }  \, \d x \, \d t \label{finite_integral}\\
    & = c_2 \,  c_1 \, C \rightarrow 0 \quad \mathrm{as} \quad c \rightarrow \infty,
\end{align}
where $C$ is the value of the integral in \eqref{finite_integral}, which is finite, since the quotient density in \eqref{finite_integral} is well defined, due to \eqref{quotient}.

Finally, we note that although the condition that the mixture components are well separated is quite strict, and might not be satisfied in practice for problems with many components, it can be relaxed to only require the components with non-zero weights in the optimal transport plan $\lambda_{ij}$ to be well-separated. The last condition is hard to check before solving the problem; however, it is satisfied for many problems in practice since the optimal transport plan $\lambda_{ij}$ is usually sparse.

\end{document}